\documentclass[twoside]{article}
\usepackage[accepted]{aistats2022}

\usepackage[utf8]{inputenc} 
\usepackage{lmodern}
\usepackage[T1]{fontenc}    
\usepackage{hyperref}       
\usepackage{url}            
\usepackage{booktabs}       
\usepackage{amsfonts}       
\usepackage{nicefrac}       
\usepackage{microtype}      
\usepackage{xcolor}         
\usepackage{textcomp}

\usepackage{amsmath}
\usepackage{graphicx}
\usepackage{floatrow}
\usepackage{pgf,tikz}
\usepackage{bm}
\usepackage{mathtools}
\usepackage{subcaption}  
\usepackage{environ}

\usepackage{multicol}
\usepackage{floatrow}
\usepackage{multirow}
\usetikzlibrary{arrows,shapes.arrows,shapes.geometric,shapes.multipart,
	decorations.pathmorphing,positioning}

\newfloatcommand{capbtabbox}{table}[][\FBwidth] 
\newcommand{\ind}{\perp\!\!\!\perp}
\NewEnviron{myequation}{%
	\begin{equation}
	\scalebox{0.9}{$\BODY$}
	\end{equation}
}
\usepackage[round]{natbib}   
\begin{document}

%

%

\twocolumn[

\aistatstitle{A Bayesian Model for Online Activity Sample Sizes}

\aistatsauthor{ Thomas S.~Richardson$^{*}$ \And Yu Liu$^{*}$ 
	\And 	James McQueen \And Doug Hains }

\aistatsaddress{University of Washington \And
Amazon.com Inc \And Amazon.com Inc \And
Amazon.com Inc} ]
\renewcommand{\thefootnote}{\fnsymbol{footnote}}
 
\def\thefootnote{*}\footnotetext{These authors contributed equally to this work} 

\renewcommand{\thefootnote}{\arabic{footnote}}
\begin{abstract}
  In many contexts it is useful to predict the number of individuals in some population who will initiate a particular activity during a given period. For example,
  the number of users who will install a software update, the number of customers who will use a new feature on a website or who will participate in an A/B test.  In practical settings, there is heterogeneity amongst individuals with regard to the distribution of time until they will initiate. 
  For these reasons it is inappropriate to assume that the number of new individuals observed on successive days will be identically distributed.  
  Given observations on the number of unique users participating in an initial period, we present a simple but novel Bayesian method for predicting the number of additional individuals who will participate during a subsequent period. We illustrate the performance of the method in predicting sample size in online experimentation.
\end{abstract}

\section{INTRODUCTION}
\label{introduction}

There are many situations in which it is necessary to predict the number of individuals who will engage in an activity for the first time during some specified period. For example, a credit card company may want to predict the number of customers who respond to an offer for a new card; online e-commerce companies may want to predict the number of registered users who opt-in to new features or upgrade their membership.  In the domain of online experimentation or A/B testing it is extremely useful to predict the number of individual users who will take part in the experiment as this can be used to estimate how long an experiment needs to run in order to get a sufficient sample size in order to achieve a desired power level. 
Such experiments are important as they provide a means by which ``end-users can help guide the development of features''  \citep{kohavi2007practical}.

The problem that we consider in this paper is as follows: given a fixed population and observations on the number of unique individuals who initiate activity during an initial period of time, for example, a week, predict the number of additional individuals who will initiate activity during a subsequent period of time, for example one or three weeks.

Though formally the data can be seen as forming a time-series, the problem is not amenable to conventional time-series models. Though time-series model like ARIMA or Dynamic Regression have been used to predict the overall level of online traffic~\citep{anderson2016library, bjorklund2021effect}, such models typically require long series and also assume stationarity of innovations after de-trending. Consequently, such methods are not suitable in many applications in which there is a wish to make predictions shortly after a new service has been launched.
The data may also be viewed as a (non-stationary) counting process for which Hawkes processes are sometimes used \citep{hawkes:1971a,hawkes:1971b}. However, since the decision of each individual regarding when (or if) they will  participate is taken independently, the process does not typically exhibit self-excitatory behavior of the type described by Hawkes processes.

Another challenge is that,  
in general, there will be heterogeneity within the population: some customers will initiate much more quickly than others; for example daily users of a mobile app will initiate much sooner than those who use it only for specific or occasional purposes. The users who are observed to initiate activity at the start of the observation period will be weighted more heavily towards frequent users than is the case subsequently. This unknown heterogeneity must be taken into account when making predictions for subsequent periods.

The primary contribution of this paper is the use of a hierarchical Bayesian Beta-Geometric model with censoring to model the heterogeneity of individuals in the population with respect to their propensity to initiate activity on any given day. Estimates for the number of additional individuals arriving in subsequent periods can then be obtained by simulating from the posterior predictive distribution. A convenient by-product of the simplicity of our model is that it is possible to sample from the posterior via straight Monte-Carlo simulation, thereby avoiding  issues that may arise with Markov Chain approaches, such as assessing convergence and mixing times, which could present difficulties in large scale or fully-automated applications.

Relating the approach described here to prior work, there is a long history of using hierarchical beta-binomial models in Bayesian statistical inference for count data; see for example, \cite{leonard:1972,novick:lewis:jackson:1973,dempster1983combining}. However, we are not aware of prior work that has applied these models to inference for the number of individuals initiating behavior in a given time period.

The rate at which customers participate in an online experiments, often called A/B tests, will have a great influence on the effect sizes that can be detected statistically within a given time period. Recent works outlining current best practice for such online experiments include \cite{kohavi2007practical} and \cite{bakshy2014designing}. However, again to the best of our knowledge, previous work has not considered the specific question of predicting how many additional unique customers will be accrued in a subsequent time interval, given the number of unique customers who participated during  an initial period. Such predictions are obviously of great relevance to decisions regarding the organization and scheduling of A/B experiments \citep{ kohavi2017online,casella2021statistical}.

In section~\ref{accrualmodel}, we describe our proposed accrual model and the assumptions underlying it. In section~\ref{inference}, we describe how to make inferences about the number of additional individuals arriving in the second period by computing the posterior predictive distribution. In section \ref{experiment} we illustrate the posterior distribution, predictive distributions and demonstrate the performance of our model on a large meta-analysis of A/B experiments.

\section{ACCRUAL MODEL}
\label{accrualmodel}

Our goal here is to build a simple accrual model which uses the number of unique users or customers who first participate  at day $t$, $t = 1, 2, \ldots, d$ to make a prediction regarding the total number of new (unique) customers who will participate in the second period of {$d^*$} days. We take $d = 7$ (first week) and  $d^*$ a multiple of 7  (multiple weeks) in our data analyses. 

\subsection{Notations and Model Assumptions}\label{subsec:notation}
Consider a set of customers indexed by $i \in \{1,\ldots ,n\}$ and ``Day $t$'' means the $t$-th day on which it was possible to participate. 

Our model makes the simplifying assumption that each customer $i$ participates for the first time on any given day with a probability $\pi_i$. We will also assume that if on day $t=1,2,\ldots$ a customer has never participated previously, 
then their probability
of participating on the next day remains $\pi_i$. Consequently, the day of the experiment on which customer $i$ first participates will follow a Geometric distribution with probability $\pi_i$.

\subsection{Simple Accrual model: $\pi$ discrete}
\label{discrete}
In the model that we propose below the  underlying heterogeneity among customers in their behavior is described by modeling $\pi_i$ as being drawn from a distribution with support on $[0,1]$.  However, to provide intuition, before describing this model, consider first the simpler setting in which $\pi_i$ is discrete, taking $100$ different uniformly spaced values: 

\[
\pi_i \in \{ 0.01,\ldots , 0.98, 0.99, 1.00\}.
\]

$\pi_i$ can be viewed as a characteristic of a customer. Thus it make sense to group together customers with the same value of $\pi$. Let $N_{\pi}$ denote the number of customers in this group. Thus, in the simple setting described, there will be   100 groups of customers. It follows that the expected number of customers of type $\pi$ who first participate on Day $1$ will be $\pi \times N_{\pi}$, the expected number who first participate on Day $2$ will be $(1-\pi)\pi N_\pi$, and the expected number participating on Day $k$ will be $(1-\pi)^{k-1}\pi N_\pi$.

Consequently, in expectation, the set of customers who participate  on Day 1 will consist of the following, ordered according to the value of $\pi$:
\[
(0.01\cdot N_{0.01},\;\;0.02\cdot N_{0.02},\ldots , 0.99\cdot N_{0.99},\;\; 1\cdot N_1).
\]
Similarly, the set of customers who first participate on Day 2 will consist, in expectation, of the following:
\begin{align*}\
(0.99\cdot 0.01\cdot N_{0.01}, \;\;0.98\cdot 0.02\cdot N_{0.02},\ldots ,\\ 0.01\cdot 0.99\cdot N_{0.99},\;\; 0\cdot 1\cdot N_1).
\end{align*}
Notice that relative to Day $1$, a smaller proportion of people with $\pi=0.99$ participate  {\it for the first time} on Day 2, since $99\%$ of such people already participated on Day $1$. By repeating this procedure, we may simulate the number and type of customers who first participate on Day $t$.

\subsection{Accrual model: Geometric likelihood with censoring}
We now consider a more realistic setting in which $\pi$ is continuous and build a model for the distribution of the participation probabilities $\pi$. Similar to section~\ref{discrete}, given the model assumptions, the day on which individual $i$ first participates will follow a Geometric distribution with probability $\pi_i$:
\begin{eqnarray}
\hbox{day on which $i$ initiates} \,|\, \pi_i &\sim&  Geo(\pi_i).
\label{eq:geom-model}
\end{eqnarray}

For each individual $i \in \{1, 2, \ldots, n\}$ who participates in the first period of length $d$ we will define $X_i$ to be the day on which they first participated;
if an individual $i$ never participates in the first period, then we define $X_i=0$.
Thus $X_i \in \{0,1,\ldots ,d\}$, where, for example, $d=7$ if the first period, for which participation data is available, is a week.
We formulate the model hierarchically, with a Beta distribution over the unknown participation probabilities. The model is specified formally as follows:
\begin{eqnarray}
p(X_i=x_i \;|\; \pi_i) &=&  \begin{cases} (1-\pi_i)^{x_i-1}\pi_i, & x_i \in \{1,\ldots ,d\};\\
(1-\pi_i)^{d}, & x_i=0;
\end{cases} \nonumber \\[4pt]
\pi_i \;|\; \alpha, \beta &\sim& \hbox{Beta}(\alpha,\beta);\nonumber   \\[4pt]
p(\alpha,\beta) &\propto& (\alpha + \beta)^{-5/2}, \quad \alpha >0, \beta >0.\label{eq:hyperprior}
\end{eqnarray}
Note that the first term here can also be rewritten as:
\begin{multline*}
p(X_i=x_i \;|\; \pi_i) = \pi_i^{\mathbb{I}(x_i>0)}(1-\pi_i)^{(x_i-1)\mathbb{I}(x_i>0)+d\,\mathbb{I}(x_i=0)}; 
\end{multline*}
Here $\propto$ indicates {\it proportional to}, while $\mathbb{I}(\cdot)$ is the indicator function that takes the value $1$ if the condition is true and $0$ otherwise.  $ p(X_i=x_i \;|\; \pi_i)$ follows a censored geometric distribution with success probability $\pi_i$,
where values larger than $d$ are censored and recorded as  $0$. We use the conjugate prior for $\pi_i$, which is Beta$(\alpha, \beta)$. The hyper prior $p(\alpha,\beta)$ is a default prior recommended by~\cite{gelmanbda04}[\S5.3].  

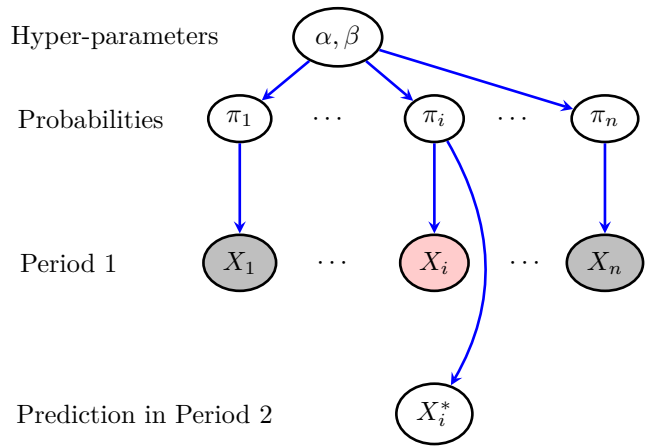
\begin{figure}
	\centering
	\begin{tikzpicture}[pre/.style={->,>=stealth,ultra thick,line width = 1pt}]
	\begin{scope}
	\node[pre,name=hpar, draw, ellipse]{$\alpha,\beta$};
	\node[left=8.5mm of hpar]{Hyper-parameters};
	\node[pre,name=pi1, below left=8mm of hpar, draw, ellipse]{$\pi_1$};
	\node[left=4.5mm of pi1]{Probabilities};
	\node[name=dots,right=4mm of pi1]{$\cdots$};
	\node[pre,name=pii, below right=8mm of hpar, draw, ellipse]{$\pi_i$};
	\node[name=dots2,right=3mm of pii]{$\cdots$};
	\node[pre,name=pin, right=14mm of pii, draw, ellipse]{$\pi_n$};
	\draw[pre,->,blue](hpar) to (pi1);
	\draw[pre,->,blue](hpar) to (pii);
	\draw[pre,->,blue](hpar) to (pin);
	\node[pre,name=x1, below=12mm of pi1, draw, ellipse, fill=lightgray]{$X_1$};
	\node[left=10mm of x1]{Period 1};
	\node[right=4mm of x1]{$\cdots$};
	\node[pre,name=xi, below=12mm of pii, draw, ellipse, fill=red!20]{$X_i$};
	\node[right=4mm of xi]{$\cdots$};
	\node[pre,name=xn, below=12mm of pin, draw, ellipse, fill=lightgray]{$X_n$};
	\draw[pre,->,blue](pi1) to (x1);
	\draw[pre,->,blue](pii) to (xi);
	\draw[pre,->,blue](pin) to (xn);
	\node[pre,name=xis, below=12mm of xi, draw, ellipse]{$X^*_i$};
	\node[left=15mm of xis]{Prediction in Period 2 };
	\draw[pre,->,blue](pii) to[out=-60,in=60] (xis);
	\end{scope}
	\end{tikzpicture}
	\caption{Graphical Model: The hierarchical Beta-Geometric model. $(\alpha,\beta)$ are hyper-parameters; for an individual $i$, $\pi_i$ is the individual-specific probability of first participating on a given day, having not participated previously; $X_i$ indicates {\it either} the day on which the customer $i$ first participates  in the first period {\it or} whether the individual did not participate in the first period ($X_i=0$). Shaded nodes are observed; nodes shaded red indicate individuals who didn't participate in the first period.
		$X^*_i$ is a prediction of when an individual who did not participate in the first period will first participate in the second period, or if they will again not participate ($X_i^*=0$).\label{fig:beta-binomial-graph1}}
\end{figure}

A graphical depiction of the proposed model is in Figure~\ref{fig:beta-binomial-graph1}. $\pi_i$ is the individual-specific probability of customer $i$ participating on a given day.  As described above, $\pi_i$ can be viewed as describing a given customer's propensity to initiate activity. Similarly,  the hyper-parameters $\alpha$, $\beta$ which determine the distribution from which the $\pi_i$ values are drawn, can be viewed as describing the character of different activities. For example, customers may be more likely to participate in an experiment changing the layout of the home page of a website than they might be to participate in an experiment changing the customer support contact page.

From the two assumptions given in Section~\ref{subsec:notation}, it follows that the parameters $(\pi_1,\ldots ,\pi_n)$ are i.i.d.~conditional on the hyper parameters $\alpha$, $\beta$.
In other words, under the model each individual's probability $\pi_i$ is an i.i.d.~draw from a Beta$(\alpha,\beta)$ distribution. 

To understand and motivate the hyper prior, we first recall the following facts about Beta distributions: 
\bigskip

If $\pi^* \sim$ Beta$(\alpha,\beta)$ then:
\begin{itemize}
	\item[(a)] $E[\pi^*] = \frac{\alpha}{\alpha+\beta} \in [0,1]$.
	\item[(b)] The variance is given by:
	\begin{equation}
	\begin{split}
	V[\pi^*] &= \frac{\alpha\beta}{(\alpha+\beta)^2(\alpha+\beta+1)}\\
	&=\frac{E[\pi^*](1-E[\pi^*])}{\alpha+\beta+1} \leq \frac{1}{4(\alpha+\beta+1)}
	\end{split}
	\end{equation}
\end{itemize}

where we have used the fact that $E[\pi^*](1-E[\pi^*]) \leq 1/4$. Consequently, for large $\alpha$, $\beta$,  sd$(\pi^*)\approx (\alpha+\beta)^{-0.5}/2$.

Consequently, the hyper-prior (\ref{eq:hyperprior}) corresponds to putting independent (improper) uniform priors on $E[\pi^*]=\alpha/(\alpha+\beta)$ and $(\alpha+\beta)^{-0.5}\approx \hbox{sd}(\pi^*)$.
Note that the hyperprior (\ref{eq:hyperprior}) used here is {\it improper}; it does not integrate to $1$. 
However, it will lead to a proper posterior provided that for at least one $i$, $0<x_i<d$; see \cite{gelmanbda04}[Ex.~5.7].

\section{INFERENCE FOR THE ACCRUAL MODEL}
~\label{inference}
In this section, we discuss how to make predictions using the proposed accrual model.
\subsection{Posterior predictive probability of continued non-participation}
~\label{inference:posterior}
We first compute the elements in the formula of the posterior predictive distribution for future point $X^*_i$ in Figure~\ref{fig:beta-binomial-graph1}. Let ${\bm x} = (x_1,\ldots ,x_n)$ be the vector recording the day of the first participation or $0$ if there was no visit during the first period for $n$ individuals, and let ${\bm \pi} = (\pi_1,\ldots, \pi_n)$ be the individual-specific probabilities.

From Bayes rule, following the independence assumptions, we can obtain the \textit{joint posterior over parameters and hyper-parameters $(\alpha,\beta)$}:
\begin{align}
\MoveEqLeft[1]{p({\bm \pi},\alpha, \beta\,|\, {\bm x})}\nonumber\\
& \propto p(\alpha,\beta) p({\bm \pi}\,|\, \alpha, \beta) p({\bm x}\,|\, {\bm \pi},\alpha,\beta)\nonumber\\
&\propto p(\alpha,\beta) \prod_{j=1}^{n} p({\pi_j}\,|\, \alpha, \beta) \prod_{k=1}^{n} p(x_k\,|\, {\pi_k})\nonumber   \\ 
&\propto p(\alpha,\beta)  \prod_{j=1}^{n} \bigg(\frac{\Gamma (\alpha + \beta)}{\Gamma(\alpha) \Gamma(\beta)}\pi_j^{\alpha-1+\mathbb{I}(x_j>0)}
\label{eq:joint-posterior} \\[-2pt]
& \quad\quad\quad\quad \times (1-\pi_j)^{\beta-1+(x_j-1)\mathbb{I}(x_j>0)+d\,\mathbb{I}(x_j=0)} \bigg).\nonumber
\end{align}
Similarly we obtain the \textit{conditional posterior distribution over $\pi_i$ given hyper-parameters $(\alpha, \beta)$ and the data $x_i$}:
\begin{myequation}
	\begin{aligned}
		&{p(\pi_i \;|\; \alpha, \beta, x_i)} \\[4pt]
		&\propto p({\pi_i}\,|\, \alpha, \beta) p({ x_i}\,|\, { \pi_i})\\[4pt]
		&= \hbox{Beta}(\alpha + \mathbb{I}(x_i>0), \beta+(x_i-1)\mathbb{I}(x_i>0)+d\,\mathbb{I}(x_i=0))\\[4pt]
		&= \frac{\Gamma (\alpha + \beta+x_i  +d\,\mathbb{I}(x_i=0))}{\Gamma(\alpha + \mathbb{I}(x_i>0)) \Gamma(\beta+ (x_i-1)\,\mathbb{I}(x_i>0) + d\,\mathbb{I}(x_i=0))}\\[4pt]
		&\qquad \times \pi_i^{\alpha -1+ \mathbb{I}(x_i>0)}(1-\pi_i)^{\beta-1+ (x_i-1)\,\mathbb{I}(x_i>0) + d\,\mathbb{I}(x_i=0)},\label{eq:posterior-for-a-single-pi}
	\end{aligned}
\end{myequation}

where we have used the fact that $\mathbb{I}(x_i>0) + (x_i-1)\,\mathbb{I}(x_i>0) = x_i\mathbb{I}(x_i>0) = x_i$.
Here (\ref{eq:posterior-for-a-single-pi})  follows from the form of the joint posterior (\ref{eq:joint-posterior}), which shows
$\pi_i \ind {\bm \pi}_{-i}, {\bm x}_{-i}\;|\; x_i, \alpha,\beta$,  where we use ${\bm \pi}_{-i} \equiv (\pi_1,\ldots ,\pi_{i-1},\pi_{i+1},\pi_n)$ to indicate the `other' elements of ${\bm \pi}$ and likewise for $ {\bm x}_{-i}$; the conditional independence here may also be obtained by applying the graphical d-separation criterion to the DAG shown in Figure \ref{fig:beta-binomial-graph1}. Thus from (\ref{eq:posterior-for-a-single-pi}) we obtain:
\begin{myequation}
	\begin{aligned}
	&{p({\bm \pi}\;|\; \alpha, \beta, {\bm x})}\\
	&=\prod_{j=1}^n  {p(\pi_i \;|\; \alpha, \beta, x_i)}\\
	&=\prod_{j=1}^n 
	\frac{\Gamma (\alpha + \beta+x_j+d\,\mathbb{I}(x_j=0))}{\Gamma(\alpha + \mathbb{I}(x_j>0)) 
			\Gamma(\beta+ (x_j-1)\,\mathbb{I}(x_j>0) + d\,\mathbb{I}(x_j=0))}\\
	&\qquad  \times \pi_j^{\alpha -1+ \mathbb{I}(x_j>0)}(1-\pi_j)^{\beta-1+ (x_j-1)\,\mathbb{I}(x_j>0) + d\,\mathbb{I}(x_j=0)}.\label{eq:pi-given-alpha-beta}
	\end{aligned}
\end{myequation}

Similar to the development in  \cite{gelmanbda04}[\S5.3, Eq.(5.5) Eq.(5.8)], the {\it posterior over the hyper-parameters $\alpha$, $\beta$} is then:
\begin{myequation}
	\begin{aligned}
		\label{eq:posterior-over-hyperparams}
		&p(\alpha, \beta \,|\, {\bm x}) \\
		&= \frac{p(\alpha,\beta,{\bm \pi}\,|\, {\bm x})}{p({\bm \pi}\,|\, \alpha,\beta,{\bm x})}\\[4pt]
		&\propto p(\alpha,\beta) \prod_{j=1}^n \biggl( \frac{\Gamma(\alpha + \beta)}{\Gamma(\alpha)\Gamma(\beta)} \\
		&\times \frac {\Gamma(\alpha + \mathbb{I}(x_j>0)) \Gamma(\beta+ (x_j-1)\,\mathbb{I}(x_j>0) + d\,\mathbb{I}(x_j=0))}
		{\Gamma (\alpha + \beta+x_j+d\,\mathbb{I}(x_j=0)}
		\biggr).
	\end{aligned}
\end{myequation}

For a given individual $i$, who did not participate in the first period, so $X_i=0$, we will define $X_i^*$ to be the first day on which the individual participates in the second period of length $d^*$,
or, as before, we define $X_i^*=0$ if they also never participate in the second period.
Given the hyper-parameters $(\alpha, \beta)$, it follows from (\ref{eq:posterior-for-a-single-pi}) that:	
\begin{myequation}
	\begin{aligned}
		\label{eq:prob-zero-for-xi-star}
		&p(X^*_i=x^*_i  \;|\; \alpha, \beta, X_i=0)\\[4pt]
		& = \int p(X^*_i=x^*_i  \;|\;  \pi_i , \alpha, \beta, X_i=0) p(\pi_i \;|\; \alpha, \beta,  X_i=0) d \pi_i\\
		& = \int p(X^*_i=x^*_i  \;|\;  \pi_i) p(\pi_i \;|\; \alpha, \beta,  X_i=0) d \pi_i\\
		& = \frac{\Gamma(\alpha + \beta + d)}{\Gamma(\alpha)\Gamma(\beta + d)} \\[4pt]
		&\times \frac{\Gamma(\alpha+\mathbb{I}(x_i^*>0))\Gamma(\beta+ d+(x_i^*-1)\mathbb{I}(x_i^*>0) + d^*\mathbb{I}(x_i^*=0))}{\Gamma(\alpha+\beta+d+x^*+d^*\mathbb{I}(x_i^*=0))}.
	\end{aligned}
\end{myequation}

Thus, conditional on hyper-parameters $(\alpha,\beta)$, the {\it probability that an individual who has not participated in the first period will again not participate in the second period} is:
\begin{equation}
\begin{aligned}
\label{eq:posterior-not-trigger-second}
\MoveEqLeft{p(X^*_i=0  \;|\; \alpha, \beta, X_i=0 )}\\
&=   \frac{\Gamma(\alpha)\Gamma(\beta+d+d^*)}{\Gamma(\alpha+\beta+d+d^*)}\frac{\Gamma(\alpha+\beta+d)}{\Gamma(\alpha)\Gamma(\beta+d)} \\
& \equiv q_{0}(\alpha,\beta,{\bm x}).
\end{aligned}
\end{equation}
We express the probability $q_{0}(\alpha,\beta,{\bm x})$ of continued non-participation as a function of the full first period data ${\bm x}$ because this determines the set of individuals $i$  who did not participate in the first period, for whom $X_i=0$.	
Note that under the model this probability will be the same for all people who did not participate in the first week.

\subsection{Predicting the number of new individuals in the second period}
Let $n_0$ be  the number of individuals who did not participate in the first period. Similarly, let $\{i_1,\ldots ,i_{n_{0}}\}$ be the subset of individuals who did not participate in the first period.

Given $\alpha$, $\beta$ the corresponding variables $\{X^*_{i_1},\ldots X^*_{i_{n_{0}}}\}$ are i.i.d., and have probability $q_0(\alpha,\beta, {\bm x})$  of taking the value $0$.
It follows that given $\alpha$, $\beta$  the {\it number} of individuals $n_{00}$ who did not participate in the first period and who again did not participate in the second period will follow a binomial distribution,
with the following parameters:\footnote{Note that the posterior distribution for $n_{00}$ is {\em not} a Beta-Binomial distribution. $n_{00}$ is Binomial given $(\alpha,\beta)$, but
	the posterior over $(\alpha,\beta)$ is {\it not} a Beta distribution.}
\begin{align}\label{eq:n00-prediction}
n_{00}\;|\; \alpha, \beta, {\bm x} \sim \hbox{Binomial}(n_{0}, q_{0}(\alpha,\beta,{\bm x}))
\end{align}

Hence a simple Monte-Carlo scheme for simulating from the posterior for $n_{00}$ is as follows:
\begin{enumerate}
	\item Simulate $\alpha$, $\beta$ from the posterior distribution (\ref{eq:posterior-over-hyperparams});
	\item For each value of $(\alpha,\beta)$ draw a number from the Binomial distribution given by (\ref{eq:n00-prediction}).
\end{enumerate}
In step one we use the (straight) Monte-Carlo Ratio-of-Uniforms algorithm~\citep{wakefield1991efficient,kinderman1977computer} to sample $\alpha$, $\beta$. A point estimator of $n_{00}$ may be obtained by taking the median of the different draws from step 2. We use the posterior median as it is more robust than the posterior mean.

Thus the number of individuals who did not participate in the first period, but did participate in the second period is:
\begin{align}
\MoveEqLeft[15]{\hbox{\it no.~of individuals participating } }\nonumber\\
\hbox{\it for first time in Period 2}&= n_0 - n_{00}.\label{eq:new-individual-identity}
\end{align}
Hence we are able to efficiently obtain samples from the posterior for the number of additional individuals participating during the second period.

\subsection{Inference when total population size is unknown}
\label{section:unknow_population}
In the development so far, we have supposed that the number of censored observations, $n_0$ corresponding to individuals who did not show up in the first period, is known. This will hold in settings where the population of users is known, for example, where subscriptions or accounts are required for participation.. However, in some situations this may not be the case.

Fortunately, this turns out not to be a practical issue: provided that the number of censored observations is large relative to the number of uncensored observations, the posterior distribution for the number of additional individuals who will participate for the first time in the second period is not sensitive to the exact number of censored observations; see Section \ref{experiment}. In more detail, one may use a plug-in estimate of $n_0$ given by a multiple $\lambda$ of the uncensored observations:
\begin{equation}\label{eq:n0}
\begin{aligned}
\hat{n}_0 = \lambda \sum_{i = 1}^{n} \mathbb{I}(x_i^*>0). 
\end{aligned}
\end{equation}
If necessary, $\lambda$ can be tuned via cross-validation~\citep{james2013introduction} on past data.

\section{Experimental Results}

\label{experiment}
In order to evaluate the performance of the model, we obtained participation data from experiments performed at Amazon.com, Inc. during the last year in the United States. We collected data from 1961 experiments that ran for at least 2 weeks,  and 976 that ran for at least 4 weeks. For each experiment we used the number of new customers first participating in the experiment each day during the first week to predict the total sample size if the experiment were to continue running for $k$ weeks in total, where $k = 2, 3, 4, 5$. (Axes in these plots are not annotated owing to business confidentiality.)

\subsection{Illustrative Example}
\begin{figure*}%
	\centering
	\begin{subfigure}{.33\columnwidth}
		\includegraphics[width=\columnwidth]{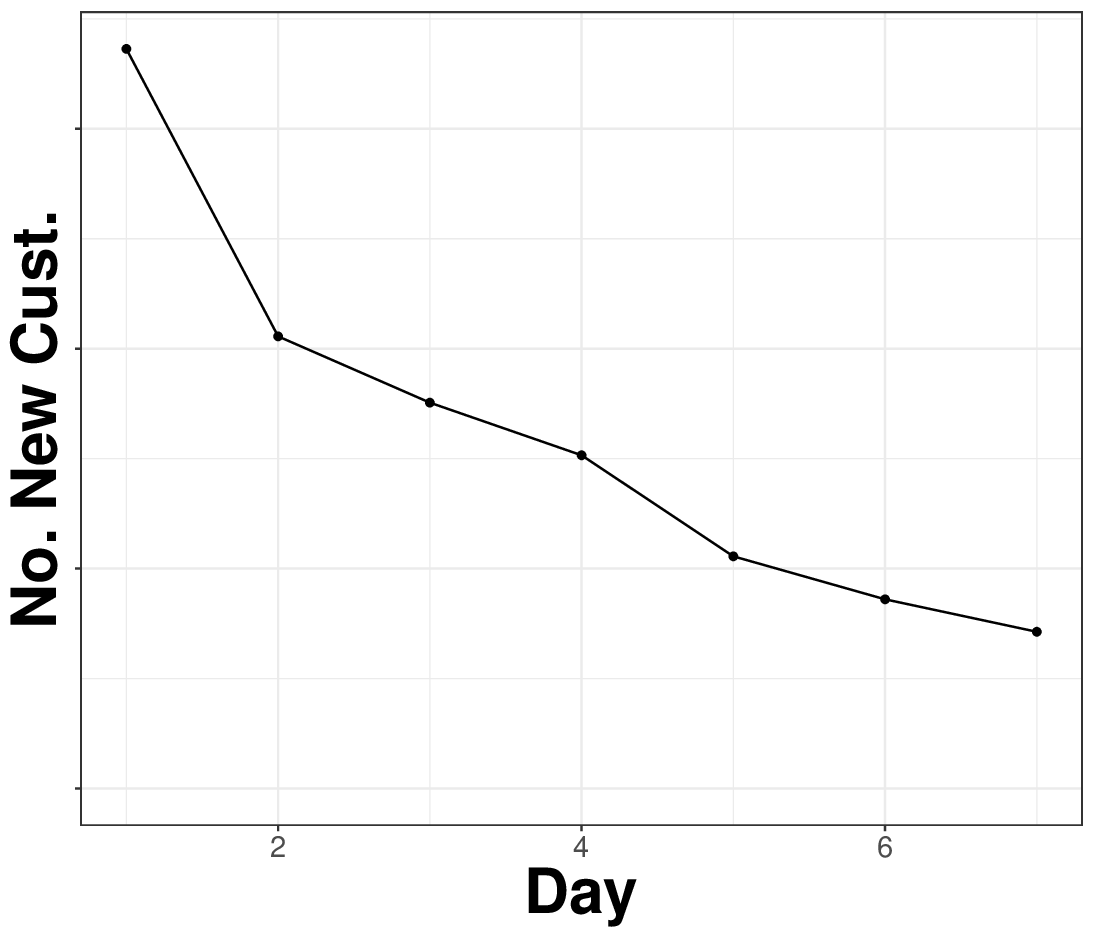}%
		\caption{Observed Data for an Experiment}%
		\label{fig:data}%
	\end{subfigure}\hfill%
	\begin{subfigure}{.32\columnwidth}
		\includegraphics[width=\columnwidth]{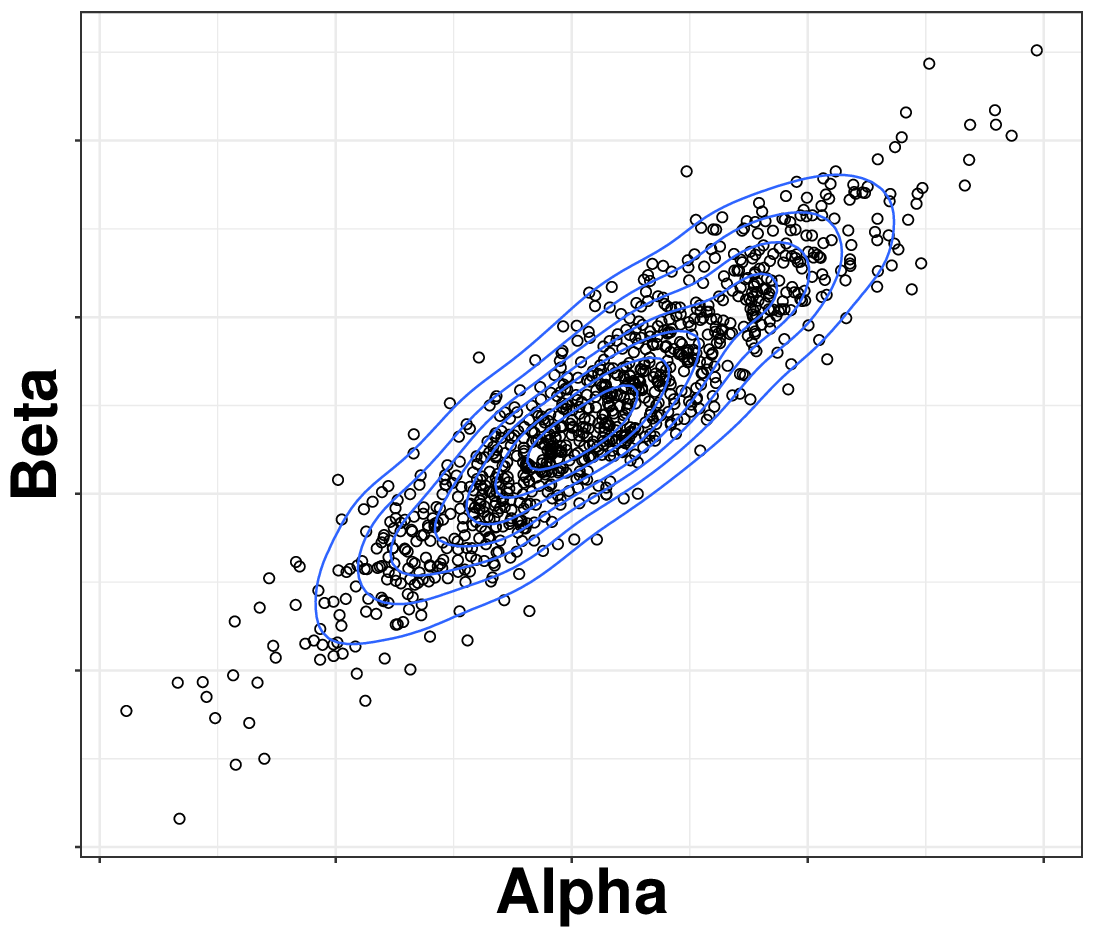}%
		\caption{Scatter Plot and Density Plot of Posterior ($\alpha$, $\beta$)} 
		\label{fig:posterior}%
	\end{subfigure}\hfill%
	\centering
	\begin{subfigure}{.33\columnwidth}
		\includegraphics[width=\columnwidth]{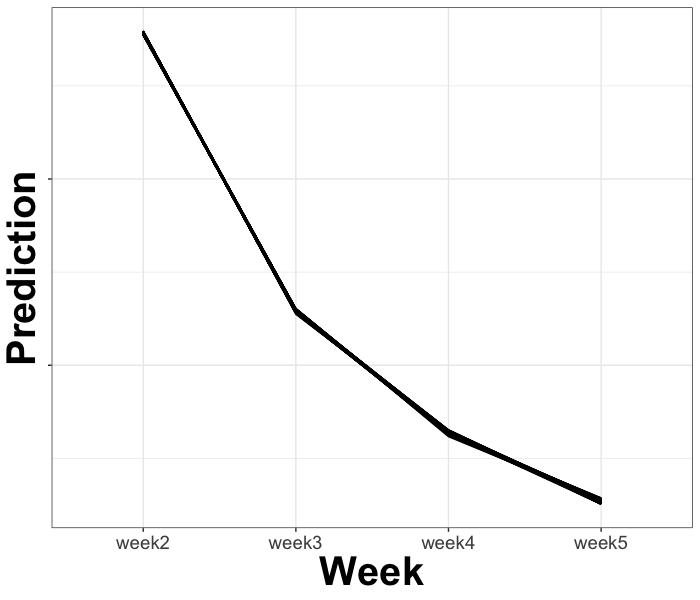}%
		\caption{Predicted Curves from 1000 Simulations.}%
		\label{fig:predicted_curve}%
	\end{subfigure}\hfill%
	\caption{Illustrative Example: For the sake of simplicity, we label customers participating in the experiment first time  as ``new customers''.  (a) the X-axis represents the number of days since the start of the experiment; the Y-axis represents the number of new customers who first participate in the experiment during each day in the initial period.  (b) Scatter plot of 1000 pairs of ($\alpha$, $\beta$) values sampled from Equation~(\ref{eq:posterior-over-hyperparams}) given the data in Figure~\ref{fig:data} with $\lambda=10$. Contour lines show the density of ($\alpha$, $\beta$)  and nested contours indicate regions of higher local density. (c) The Y-axis represents the predicted number of new customers participating in the experiment at week $k$, where $k \in\{1, 2, 3, 4\}$. The plot displays 1000 posterior draws for the prediction functions corresponding to the  ($\alpha$, $\beta$) samples in Figure~\ref{fig:data}. 
	}
	\label{Figure 2}
\end{figure*}

We first demonstrate the method using data from a single experiment that ran for at least 2 weeks. 
Figure~\ref{fig:data}~ shows the participation data from the first week of the experiment. Here the X-axis represents the day index $t ={1, 2, \ldots, 7}$, while the Y-axis is the number of customers who first participate in the experiment on day $t$. 

Following the development in Section \ref{inference} for this experiment we first obtained samples from the joint posterior distribution over the hyper-parameters $\alpha$ and $\beta$, and then drew samples from the joint posterior predictive distribution for the  number of new customers who first participate in each successive week.  We define $S_t$  to be the number of new customers who first participate in the experiment on day $t$. Thus the vector ${\bm x} = (x_1,\ldots ,x_n)$ in Section~\ref{inference:posterior} will consist of  $S_1$ ones, $S_2$ twos \ldots , $S_7$ sevens. Since data on the total population size was not available, the number of individuals who did not participate in the first week was estimated via (\ref{eq:n0}) with $\lambda=10$; note that this may be expressed as $ \hat{n}_0=\lambda \sum_{i = 1}^{7} S_i$ and is an estimate of the number of zeros in the vector ${\bm x}$. Here $\lambda=10$ was selected by tuning from a set of four past experiments. Since $n$ can be large in some experiments, the use of the sufficient statistics ($S_1,\ldots , S_7)$ can greatly speed up the computation relative to simply using ${\bm x}$. 


For the experimental data shown in Figure~\ref{fig:data}, Figure~\ref{fig:posterior} shows a scatterplot of 1000 samples of the hyper-parameters  $(\alpha,\beta)$  drawn from posterior distribution~(\ref{eq:posterior-over-hyperparams}) together with an estimated density plot. 
We used an implementation of the (straight) Monte-Carlo Ratio-of-Uniforms algorithm~\citep{wakefield1991efficient,kinderman1977computer,package:rust,package:bang} to obtain these samples. An implementation using sufficient statistics, based on code from the Bang package~\citep{package:bang}, can be found in github~\citep{package:proposed}. The actual scales of the X and Y axes in Figure~\ref{fig:posterior} are small indicating that, under the model, the posterior distribution for $\alpha$ and $\beta$ is concentrated.

\begin{figure*}%
	
	\begin{subfigure}{.45\columnwidth}
		\includegraphics[width=\columnwidth]{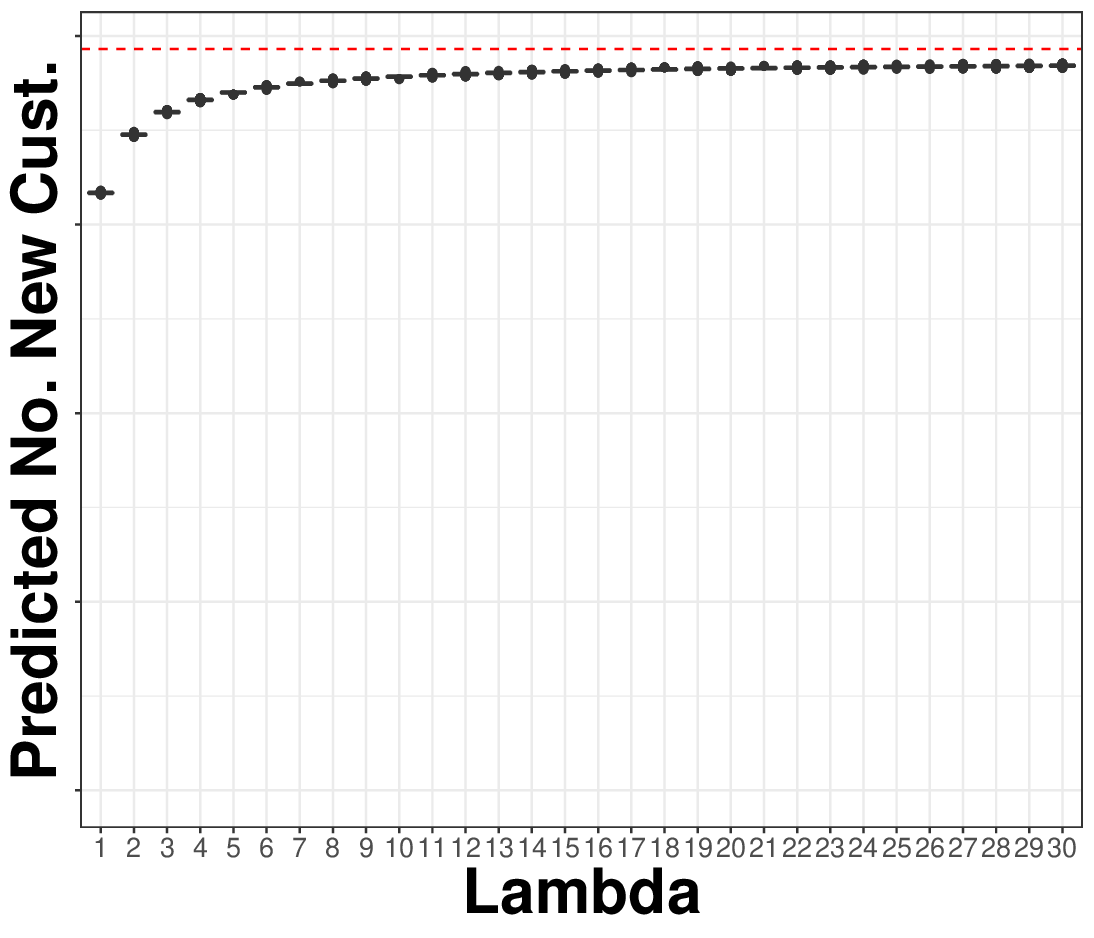}%
		\caption{Box Plots of the Predicted Number of New Customers Participating in the Experiment at Week 2 Against Different $\lambda$ Values }%
		\label{fig:boxplot}%
	\end{subfigure}\hfill%
	\begin{subfigure}{.5\columnwidth}
		\includegraphics[width=\columnwidth]{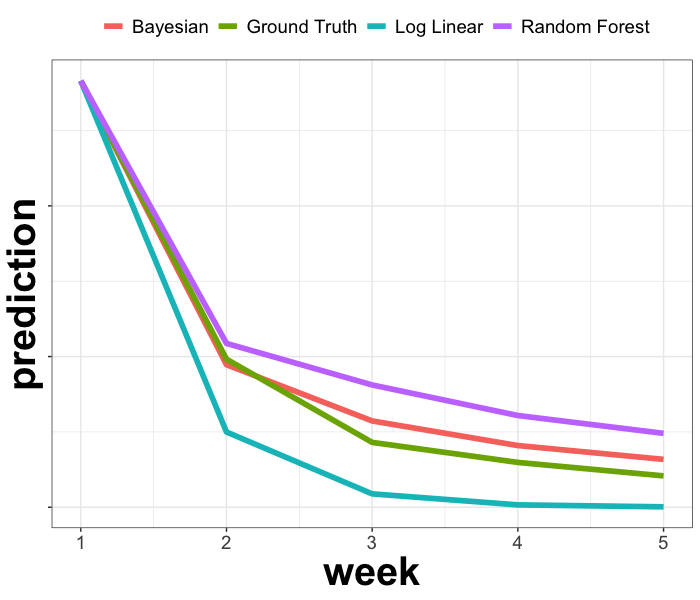}%
		\caption{Comparison of Predictions Using Different Methods }%
		\label{fig:compare}%
	\end{subfigure}\hfill%
	\caption{Results For Illustrative Example: (a) Box plots of the predicted number of new customers participating in the experiment in Figure~\ref{fig:data} at week 2  using different lambda values, $\lambda\in \{1, 2, \ldots ,30\}$. The X-axis is the value of lambda. Left bottom corner of the plot is $(0, 0)$.  Each boxplot displays the distribution of 1000 draws from the posterior predictive distribution for the number of new customers at week 2 obtained when using a specific value of lambda. The Red dotted line is the ground truth observed at week 2. (b) Comparison of the predicted number of new customers participating in the experiment shown in Figure~\ref{fig:data} at the $k$-th week using different methods. The X-axis represents the week index and the Y-axis represents the number of customers first participating in the experiment at week $k$. The true number for week 1 is observed; the numbers for weeks 2, 3, 4, 5 are predicted using different methods. The red curve shows the predictions using the proposed Bayesian model. Predictions from the Log Linear and Random Forest models are given by the blue and purple curves respectively. The ground truth is given by the green curve.
	} 
	\label{Figure 3}
\end{figure*}

For each pair of  $(\alpha,\beta)$ values, we then use Equation~(\ref{eq:posterior-not-trigger-second}) with $d^* = \{7, 14, 21, 28\}$ to obtain samples from the joint posterior predictive distribution for the probability that an individual who did not participate in the experiment in the first week will again not participate in the subsequent 1, 2, 3 and 4 weeks if the experiment were to continue running.  Plugging these probabilities and $\hat{n}_0$ into Equation~(\ref{eq:n00-prediction}) results in predictions for the number of customers who did not participate in the initial period and who also do not participate  in the second period of $d^*$ days, denoted $\hat{n}_{00}(d^*)$ where $d^* = \{7, 14, 21, 28\}$. Thus the predicted number of customers first participating in the experiment in the $k$-th week is $ \hat{n}_{00}((k-2)*7) -  \hat{n}_{00}((k-1)*7)$ where $k \in \{2, 3, 4, 5\}$ and $\hat{n}_{00}(0) = \hat{n}_0$. We repeat this procedure for all  pairs of $\alpha$ and $\beta$ values.  This results in samples from the posterior distribution over the function giving the number of new customers in each subsequent successive week. The 1000 predicted curves corresponding to the posterior draws are shown in Figure~\ref{fig:predicted_curve}.  The predicted curves show little variability, 
which is perhaps not surprising, given that the posterior over $\alpha$ and $\beta$, shown in Figure~\ref{fig:posterior}, is highly concentrated,

In Figure~\ref{fig:boxplot} we show how the predicted number of customers first participating in the experiment at week 2 is relatively insensitive to changes in $\lambda$.  Specifically, Figure~\ref{fig:boxplot} shows  boxplots giving 1000 posterior draws for each value of $\lambda \in \{1, 2, 3\ldots, 30 \}$. 
As can be seen, when $\lambda$ ranges from $8$ to $30$, the median of the boxplot for the posterior predictive distribution is approximately constant. 
The red dotted line in this plot is the ground truth.  Though the posterior median underestimates the ground truth, the accuracy is adequate for practical purposes.
\begin{figure*}[!t]%
	\centering
	\begin{subfigure}{.45\columnwidth}
		\includegraphics[width=\columnwidth]{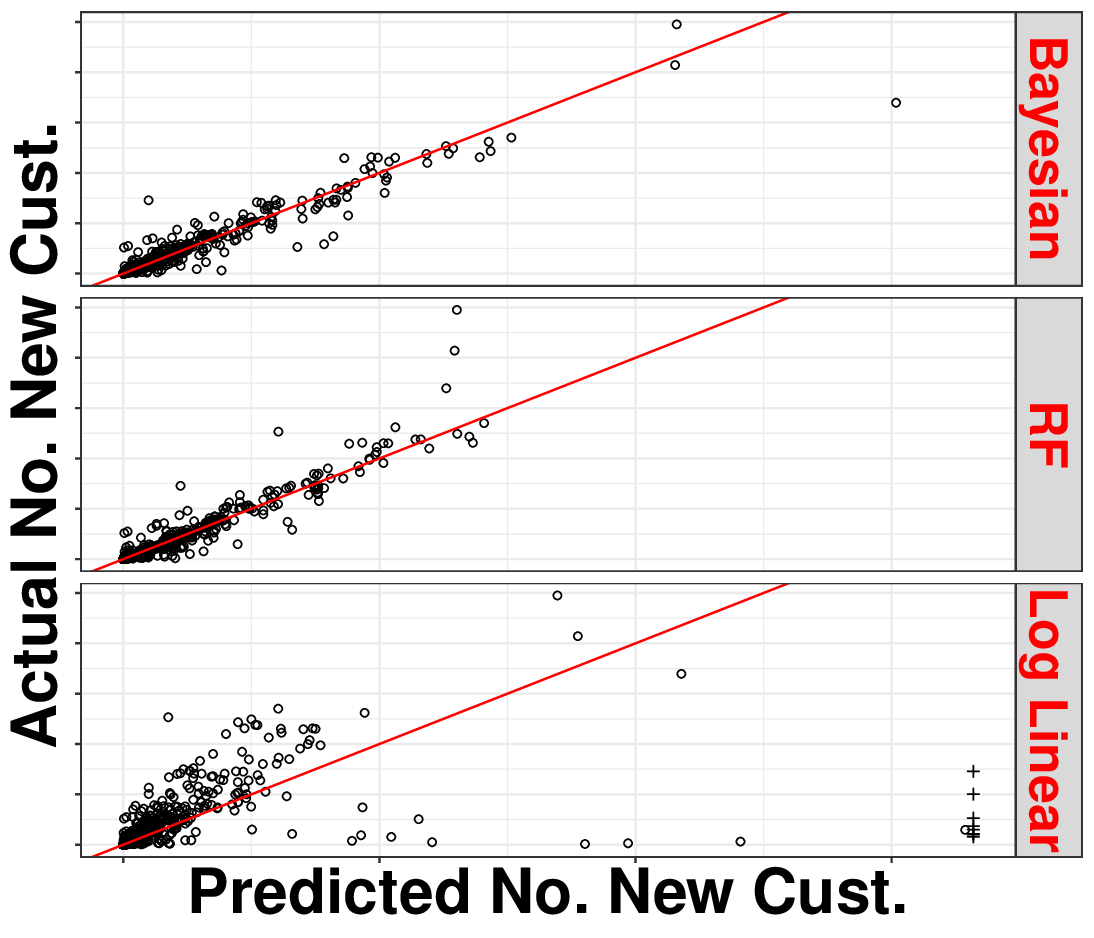}
		\caption{Comparison at Week 2}%
		\label{fig:week2}%
	\end{subfigure}\hfill%
	\begin{subfigure}{.45\columnwidth}
		\includegraphics[width=\columnwidth]{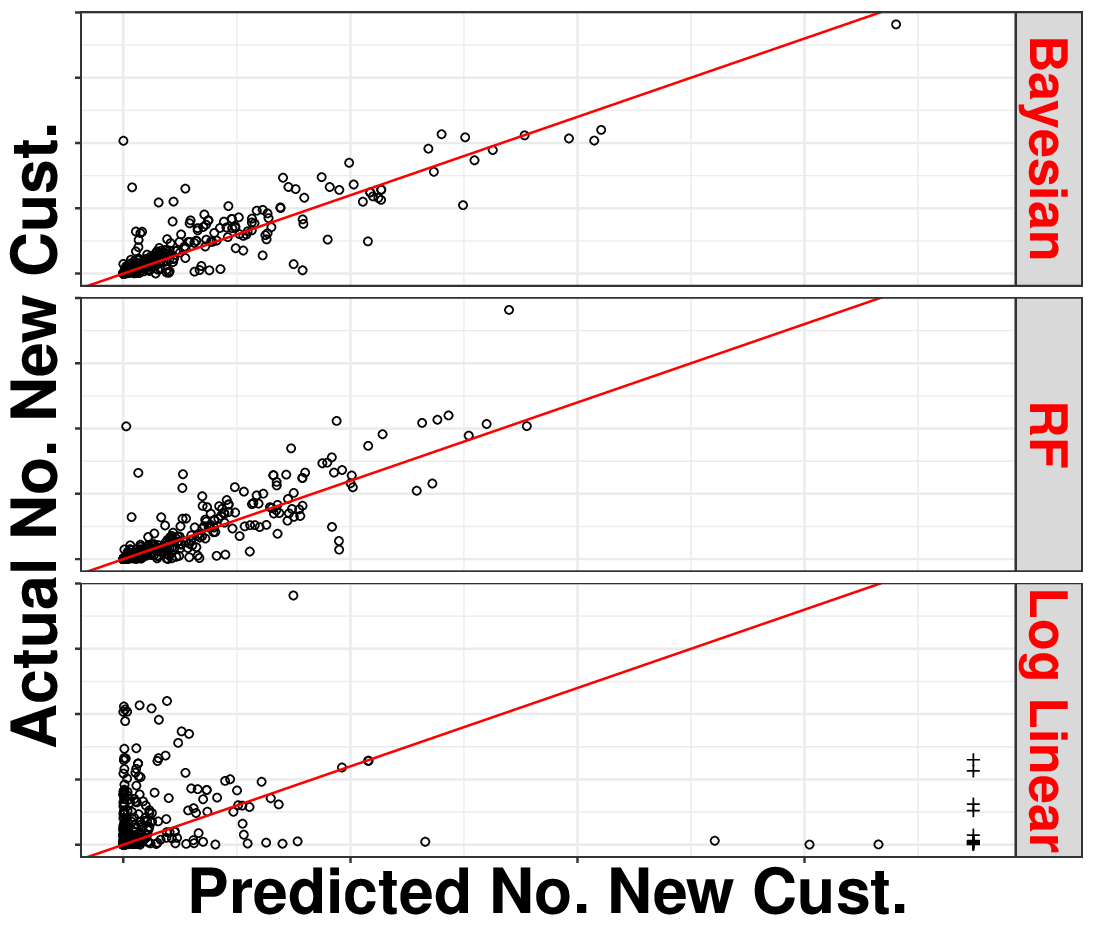}
		\caption{Comparison at Week 4}
		\label{fig:week4}%
	\end{subfigure}\hfill%
	\caption{Meta-Analysis Results: The vertical axes show the true number of new customers participating in an experiment at the $k$-th week; the horizontal axes give the predicted count of new customers participating based on the first week of data. The $45^\circ$ line, corresponding to perfect prediction, is shown in red. 
		Scatter plots contain predictions at week $k$ for experiments running for at least $k$ weeks. 
		The left panel shows results for $981$ experiments that ran for $2$ weeks; the right panel shows $488$ experiments that ran for $4$ weeks.
		The plots show the performance of the proposed predictor (top),  random forest regression (middle) and the Log Linear predictor (bottom)  at (a) 2 weeks and (b) 4 weeks.  For  visualization purposes, outliers in the Log Linear predictions are Winsorized at the maximum value on the horizontal axis and indicated by $+$.}    
	
	\label{Figure 5}
\end{figure*}

\subsection{Comparison of Bayesian Model to Other Approaches}
We compared our approach to the following three baseline models: 1. Time-Series Models;  2. Log-Linear Models; 3. Random Forest Models. 
Results from two further baselines are given in the Supplementary Materials.

Time-series (ARIMA) models are widely used to predict online traffic trends~\citep{anderson2016library,bjorklund2021effect}. However, there
are several obstacles to applying such models in our context:  (1) online A/B tests often take place during a relatively short experiment window and thus there are  an insufficient number of observed time points; 
(2) The participation data are non-stationary requiring the specification of a model for the trend, but the trend here is of primary interest! 
(3) Though in some other settings it is possible to model the trend as a function of relevant baseline covariates, these may not be available.
 We fitted ARIMA models to the initial 7 days of data using {\tt auto.arima} in the  {\tt forecast} R package~\citep{R:forecast}. Random walk ARIMA(0, 1, 0) was fitted but generated very poor predictions. We also tried to de-trend using linear regression on the number of days $t$, but this gives poor predictions due to the extrapolation. Since the time-series approaches were qualitatively much less competitive than the others, we do not consider these models further here.

Instead, we fitted a simple log-linear model: 
\begin{equation}
\log (T_d+1) =\beta_0 + \beta_1 d + \epsilon_d,
\end{equation}
where $T_d$ is the number of customers who first participate in the experiment on day $d$  and $\epsilon_d$ are independent errors with $E[\epsilon_d \,|\, d]=0$. Estimates for $\beta_0$ and $\beta_1$ may be obtained by  linear regression of the log of the number of new participants on day $d$ against $d$, for $d=1,\ldots ,7$.
The predicted number of new customers at week $k$ can be extrapolated by plugging in $d=7*(k-1)+1,\ldots, 7*k$ and then summing the predictions.

A Random Forest Model~\citep{breiman2001random} is a flexible and general machine learning model. It uses bagging and ensemble techniques to obtain estimates from a set of regression trees. Here we assume that there is a training set of previous participation data, both from the initial period {\em and} the subsequent periods for a large set of experiments. In more detail, the $i$-th row and $j$-th column of the feature matrix $\mathbf{S}$ is $S_{i, j}$: the number of customers first participating in the experiment $i$ at day $j$ where $j \in \{1,\ldots ,7\}$. Let the response variable be $F_{i, k}$, the number of customers first participating in the experiment $i$ at week $k$, where $k \in \{2, 3, 4, 5\}$. We fit 4 different random forest models on the training set: $F_k \sim RF(\mathbf{S})$, where $k = 2, 3, 4, 5$ and make prediction for each model on the test set. We use the random forest package in R~\citep{package:rf} with 500 trees and default number of variables ($2-3$) randomly sampled as candidates at each split as parameters. 

\subsection{Predictive Performance}

Figure~\ref{fig:compare} compares the predicted curves using the proposed Bayesian model, log linear regression, random forest regression and the ground truth for the example experiment in Figure~\ref{fig:data}. We use the median among 1000 simulations as the predictor for the proposed Bayesian model. For the random forest, the test data is  the feature matrix of the target experiment while the training data are the feature matrix and responses for experiments (approximately 300) that took place before the target experiment. The selection of training data here mimics the way that such a model would be implemented in practice. As illustrated in Figure~\ref{fig:compare}, the Bayesian predictor has the best performance in this illustrative example.

To assess performance on the full meta-analysis, we sort the 1961 experiments that ran for at least 2 weeks by their start date and, in order to accommodate the Random Forest, we split the first half of this set as the training dataset and the second half as the testing dataset. As before, we used $\lambda = 10$ in the Bayesian model.  We repeat the same procedure for 976 experiments running for at least 4 weeks and predict the number of customers first participating at week 4. Figure~\ref{Figure 5} shows the true number of new customers participating in the experiment at the $k$th week vs the predicted values using different methods. The $y = x$ line is shown in red in order to aid visual inspection of the performance of the various models. The plots suggest that (1)  the Bayesian method has similar performance to the random forest model while the log linear predictor is unstable; (2) The Bayesian model tends to underestimate the ground truth.

We use the  Root Mean Square Error (RMSE) and Mean absolute percentage error (MAPE) to quantify the predictive performance. 
\[
RMSE = \sqrt{\sum_{i= 1}^{n} \frac{(Predicted_i - Actual_i)^2}{n}}.
\]
\[
MAPE = \frac{100}{n}\sum_{i= 1}^{n} \left|\frac{Predicted_i - Actual_i}{Actual_i}\right|.
\]
Table~\ref{table:forecast_performance} compares the RMSE and MAPE for the random forest regression, log linear regression and the proposed Bayesian estimator. These results confirm that the Bayesian model slightly outperforms the log-linear and random forest model in terms of MAPE and has similar performance to the random forest model in terms of RMSE. Both models outperform the Log Linear model. Note that whereas the Random Forest model has access to training data, this is not used by the log-linear or Bayesian approaches (other than the four experiments that were used for selecting $\lambda$).
\begin{table}[!htbp]
	\caption[Optional captions]{Meta-Analysis Predictive Performance: RMSE and MAPE comparison between the Log Linear, Random Forest regression and Bayesian predictors.}
	\label{table:forecast_performance}
	\centering
	\small
	\begin{tabular}{|@{\,}c@{\,}|c | c |c|  c|}
		\hline
		\multicolumn{1}{|c}{} & \multicolumn{2}{c|}{Week 2} & \multicolumn{2}{c|}{Week 4} \\
		\multicolumn{1}{|c}{}&  \multicolumn{1}{c}{RMSE}& \multicolumn{1}{c|}{MAPE} &  \multicolumn{1}{c}{RMSE}& \multicolumn{1}{c|}{MAPE} \\
		\hline
		Log-Lin &1.84e+11  & 5.72e+5\% &9.06e+17  & 1.12e+13\%\\
		\hline
		RF & 7.05e+05&44.06\% &5.49e+05 & 165.22\%\\
		\hline	
		Bayes& 7.11e+05&32.59\% & 5.01e+05& 84.57\%\\
		\hline
	\end{tabular}
\end{table}

In Appendix~\ref{appendix:a} we include additional experiments comparing the performance of the proposed method with a Neural Net model and a Censored Weibull model. These results confirm that the proposed method is competitive with the other methods. 

Note that when $n_0$ is known the proposed Bayesian method does not require any past experiments as training data and thus may be applied in contexts where past experiments are either not available or have a different distribution compared to the target experiment; even when $n_0$ is unknown, only a small number of experiments are typically required to tune $\lambda$.  In contrast, traditional machine learning methods, require training data from the same distribution. The latter assumption may not hold if experiments are conducted by different teams or at different times. Indeed, this may explain some of the results given in the Supplement, where the proposed Bayesian method outperforms the Random Forest in Table~\ref{table:additional_experiment}.
Compared with the Log Linear and Censored Weibull methods, the proposed method is derived from a simple, interpretable model of customer behavior.

\section{Conclusion}
We have proposed a Bayesian approach to predict online activity in a fixed population. We demonstrated the utility of this method by applying it to predict sample size in online A/B testing. We evaluated the performance of our predictions by comparing them to the ground truth and other baselines in a large collection of  online experiments. Our results show the practical utility of the proposed method. 

The proposed Bayesian method does not require past experiments to use as training data when $n_0$ is known.
It also does not require stationarity assumptions, or long observation periods. 
The proposed method also takes into consideration the heterogeneity within the population and thus captures differences in customer behavior when using online services. Our method is simple but effective, and is being used in production to predict sample sizes for online experiment in Amazon.

There are several further refinements that we intend to explore. As indicated by Figure~\ref{fig:boxplot}, we found that in practice the proposed model typically slightly underestimates the ground truth.  One possible explanation is that there may be growth in the pool of potential users that is occurring while the experiments is running. This suggests that it may be possible to improve predictions by incorporating additional (dynamic) information regarding population size.

\clearpage
\bibliography{ref}


\clearpage
\appendix

\thispagestyle{empty}

\onecolumn \makesupplementtitle

\section{ADDITIONAL EXPERIMENTS}
\label{appendix:a}
Prompted by a suggestion from the reviewers, we added two additional baseline models for comparison, a Neural Network model and a Censored Weibull model. In addition, we compared the performance of all these methods on a recent public dataset containing online experiments \citep{liu2021datasets};
among the 78 experiments in this set, 10 (respectively, 8) experiments contain complete daily sample sizes from day $1$ to $14$ (respectively, 28).  We show the results of predictions on these experiments in Table~\ref{table:additional_experiment};  given the small size of this public dataset, the Random Forest and Neural Net models were trained on the proprietary data.

In more detail, the Censored Weibull model \citep{kay1977proportional} was fitted via maximum likelihood with data from week 1 (plus ${n_0}$ estimated by (\ref{eq:n0})).
The Neural Network \citep{neuralnetmodel} used Relu as activation function and 3 hidden layers with 16, 32 and 16 nodes respectively; $10\%$ training data was used as a validation dataset to tune other parameters.  

The results from Table~\ref{table:additional_experiment} confirm that the proposed Bayesian model is competitive against all the other methods considered. We also note here that the Neural Net  method requires greater computational resources than the other methods considered and hence is less scalable. 

Lastly, following another suggestion from the reviewers, in the last two rows of the Table we compare results for the posterior mean and median of the Bayesian method. In our experiment they are statistically indistinguishable.

\begin{table}[!htbp]
	\caption{Additional Results On Predictive Performance: RMSE and MAPE for the Log Linear, RF\,(random forest), NN\,(neural network), Censored Weibull and the proposed Bayesian method on public and proprietary data. {\bf Bold} (\underline{underline}) are best (second best). 
		For Log Linear, Censored Weibull and Bayesian,  we use the number of new samples for each day in week 1 to train and forecast the number of new samples in week 2 and week 4 (Weibull and Bayesian used $\lambda =10$). Due to the small size of the public dataset, for RF and NN, proprietary data were used to train the models.} 
		\renewcommand{\arraystretch}{1.3}
	\label{table:additional_experiment}
	\centering
	\small
	\begin{tabular}{|c |c | c |c|  c|c | c |c|  c|}
	
		\hline
		\multirow{2}{*}{Method}	& \multicolumn{4}{c|}{\multirow{2}{*}{Public Data}} & \multicolumn{4}{c|}{Proprietary data used in paper (additional } \\[-2pt]
			& \multicolumn{4}{c|}{} & \multicolumn{4}{c|}{results for NN and Weibull)}\\
		\hline
		& \multicolumn{2}{c|}{Week 2} & \multicolumn{2}{c|}{Week 4} & \multicolumn{2}{c|}{Week 2} & \multicolumn{2}{c|}{Week 4} \\
		\hline
		 & RMSE&MAPE & RMSE& MAPE & RMSE&MAPE & RMSE& MAPE\\
		\hline
		RF + & \multirow{2}{*}{6.48e+5} & \multirow{2}{*}{22.78\%}&\multirow{2}{*}{1.79e+6} & 
		\multirow{2}{*}{\underline{50.43\%}}& \multirow{2}{*}{\underline{7.05e+05}} & \multirow{2}{*}{\underline{44.06\%}} & \multirow{2}{*}{\underline{5.49e+05}}& \multirow{2}{*}{165.22\%} \\[-2pt]
		\textbf{training data}  & & & & & & & &\\
		\hline
		NN + & \multirow{2}{*}{ 1.10e+6} &  \multirow{2}{*}{47.58\%}& \multirow{2}{*}{1.92e+6} &  \multirow{2}{*}{115.86\%}&  \multirow{2}{*}{\bf{5.64e+05}}&  \multirow{2}{*}{4.65e+3\%}&  \multirow{2}{*}{5.81e+05}&  \multirow{2}{*}{490.41\%}\\[-2pt]
		\textbf{training data}  & & & & & & & &\\
		\hline	
		 \multirow{2}{*}{Log-linear} & \multirow{2}{*}{\bf{1.12e+5}}  &  \multirow{2}{*}{\underline{19.06\%}} & \multirow{2}{*}{6.86e+5}  &  \multirow{2}{*}{ 67.93\% }&  \multirow{2}{*}{1.84e+11}  & \multirow{2}{*}{ 5.72e+5\%} & \multirow{2}{*}{9.06e+17}  &  \multirow{2}{*}{1.12e+13\%}\\
		 & & & & & & & &\\
		\hline
		 Censored  & \multirow{2}{*}{4.36e+5}  & \multirow{2}{*}{32.42\%}&  \multirow{2}{*}{\bf{3.09e+5}}& \multirow{2}{*}{70.97\%} & \multirow{2}{*}{1.85e+06} &  \multirow{2}{*}{49.17\%} &  \multirow{2}{*}{1.41e+06} &  \multirow{2}{*}{\underline{139.68\%}} \\[-2pt]
		 Weibull& & & & & & & &\\
		\hline
		 Bayesian& \multirow{2}{*}{1.59e+5} & \multirow{2}{*}{12.79\%} & \multirow{2}{*}{5.09e+5}& \multirow{2}{*}{15.24\%} & \multirow{2}{*}{7.11e+05}&\multirow{2}{*}{32.57\%} & \multirow{2}{*}{5.01e+05} & \multirow{2}{*}{84.55\%} \\[-2pt]
		(mean)& & & & & & & &\\
		\hline
		 Bayesian & \multirow{2}{*}{\underline{1.59e+5}}& \multirow{2}{*}{\bf{12.78}\%} &  \multirow{2}{*}{\underline{5.09e+5}} &\multirow{2}{*}{\bf{15.25}\%} & \multirow{2}{*}{7.11e+05}&\multirow{2}{*}{\bf{32.59}\%} & \multirow{2}{*}{\bf{5.01e+05}}& \multirow{2}{*}{\bf{84.57}\%} \\[-2pt]
		(median)& & & & & & & &\\
		\hline	
	\end{tabular}
\end{table}
\vspace{-2mm}



\end{document}